\documentclass{article}

    \PassOptionsToPackage{numbers, compress}{natbib}
\usepackage[preprint]{neurips_2022}

\usepackage[utf8]{inputenc} % allow utf-8 input
\usepackage[T1]{fontenc}    % use 8-bit T1 fonts
\usepackage{hyperref}       % hyperlinks
\usepackage{url}            % simple URL typesetting
\usepackage{booktabs}       % professional-quality tables
\usepackage{amsfonts}       % blackboard math symbols
\usepackage{nicefrac}       % compact symbols for 1/2, etc.
\usepackage{microtype}      % microtypography
\usepackage{xcolor}         % colors
\usepackage{graphicx}
\usepackage{amsmath}
\usepackage{esvect}
\usepackage{tikz}
\usepackage{mwe}
\usepackage{colortbl}

\tikzset{
     block/.style={rectangle, draw, fill=red!40, text width=6em,
                   text centered, rounded corners, minimum height=3em},
     arrow/.style={-{Stealth[]}}
     }

\title{VNT-Net: Rotational Invariant Vector Neuron Transformers}

\author{%
  Hedi Zisling \\
  Department of Computer Science\\
  Ben-Gurion University of the Negev\\
  Beer-Sheva, Israel \\
  \texttt{hediz@post.bgu.ac.il} \\
  \And
  Andrei Sharf \\
  Department of Computer Science\\
  Ben-Gurion University of the Negev\\
  Beer-Sheva, Israel \\
  \texttt{asharf@bgu.ac.il} \\
}

\begin{document}

\maketitle

%%
%% The "title" command has an optional parameter,
%% allowing the author to define a "short title" to be used in page headers.

%%
%% The "author" command and its associated commands are used to define
%% the authors and their affiliations.
%% Of note is the shared affiliation of the first two authors, and the
%% "authornote" and "authornotemark" commands
%% used to denote shared contribution to the research.

%%
%% By default, the full list of authors will be used in the page
%% headers. Often, this list is too long, and will overlap
%% other information printed in the page headers. This command allows
%% the author to define a more concise list
%% of authors' names for this purpose.

%%
%% The abstract is a short summary of the work to be presented in the
%% article.
\begin{abstract}

Learning 3D point sets with rotational invariance is an important and challenging problem in machine learning. Through rotational invariant architectures, 3D point cloud neural networks are relieved from requiring a canonical global pose and from exhaustive data augmentation with all possible rotations.
In this work we introduce a rotational invariant neural network by combining recently introduced vector neurons with self attention layers to build a point cloud vector neuron transformer network (VNT-Net).
Vector neurons are known for their simplicity and versatility in representing SO(3) actions and thereby incorporated in common neural operations.
Similarly, Transformer architectures have gained popularity and recently were showed successful for images by applying directly on sequences of image patches and achieving superior performance and convergence.
% VIT took mach more resources then convolutional by alot but in point cloud the Transformer with allot of changes could be comparable to convolutional computational resources.
In order to benefit from both worlds we combine the two structures by mainly showing how to adapt the multi-headed attention layers to comply with vector neurons operations.
Through this adaptation attention layers become SO(3) and the overall network becomes rotational invariant.
Experiments demonstrate that our network efficiently handles 3D point cloud objects in arbitrary poses. We also show that our network achieves higher accuracy when compared to related state-of-the-art methods and requires less training due to a smaller number of hyper parameters in common classification and segmentation tasks .

\end{abstract}

%%
%% The code below is generated by the tool at http://dl.acm.org/ccs.cfm.
%% Please copy and paste the code instead of the example below.
%%

%%
%% Keywords. The author(s) should pick words that accurately describe
%% the work being presented. Separate the keywords with commas.
%% A "teaser" image appears between the author and affiliation
%% information and the body of the document, and typically spans the
%% page.

%%
%% This command processes the author and affiliation and title
%% information and builds the first part of the formatted document.

\section{Introduction}
%%%%%%%%%%%%%% Vector Neurons https://arxiv.org/abs/2104.12229

Unlike 2D images, point clouds lack an underlying regular structure, making
it challenging to design neural networks to process them.
Inspired by the progress of convolutional neural networks (CNNs) in the field
of image processing, many recent point cloud networks aim at defining convolution operators that can handle unordered point sets and aggregate local features by incorporating order-invariant and/or order-equivariant layers such as the multi-layer perceptron (MLP) of~\citet{qi2017pointnet} and others~\cite{li2018pointcnn,atzmon2018point,sun2020acne}.

%These methods either reorder the input point sequence or voxelize the point cloud to obtain a canonical domain for convolutions.

While order invariance has been widely explored for point cloud neural networks, rotational invariance has been just starting to gain popularity. Nonetheless, it is crucial that point cloud neural networks are relieved from requiring all objects to be aligned in one canonical pose and from exhaustive data augmentation with all possible rotations. 

For this purpose, there is a need for network layers that are equivariant to both order and SO(3) symmetries.
Recently, two approaches have been introduced to tackle this: Tensor Field Networks~\cite{thomas2018tensor}  and SE(3)-Transformers ~\cite{fuchs2020se}. While guaranteeing equivariance, both frameworks involve an intricate formulation and are hard to incorporate into existing pipelines as they are restricted to convolutions and rely on relative positions of adjacent points.
To address these issues, a simple, lightweight framework to build SO(3) equivariant
and invariant pointcloud networks was introduced based on Vector Neurons ~\cite{deng2021vn}. Their method defines a latent matrix
representations which supports direct mapping of rotations from the input pointcloud to intermediate layers. 3D generalization of classical activation functions enables prediction of activation directions which in turn guarantee equivariant layers.

Inspired by the simplicity and efficiency of vector neurons, we take another step further in this path and show how to incorporate them into self attention layers (Transformers).
Transformer networks have demonstrated promising results in natural
language processing~\cite{vaswani2017}, images~\cite{DosovitskiyB0WZ21} and point cloud ~\cite{2021PCT} tasks. In general, these models are superior in quality while being more parallelizable and requiring significantly less time to train.

\begin{figure*} [t]
    \centering
    \includegraphics[width=\linewidth]{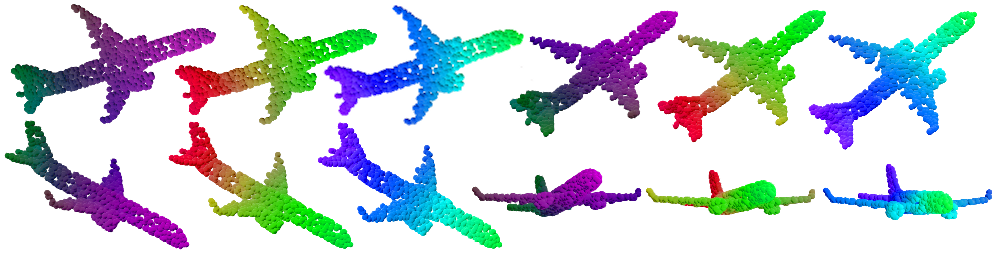}
    \caption{\textbf{VNAttention} layer yields a rotational invariant attention matrix. 3D airplanes in different orientations are invariant to our network and thus colored the same by different attention matrix weights.}    
    \label{fig:phyDnet_arch}
\end{figure*}

In this work, we incorporate vector neurons into self attention layers to build a point cloud transformer network that is rotational invariant (see Figure~\ref{fig:phyDnet_arch}).
The core idea is that self attention layers typically perform on scalars. Instead, we adapt self attention layers to handle vector neural representation which enable SO(3) rotational equivariance.
Thus, our vector neurons transformers network (VNT-Net) is rotational invariant. Furthermore it is both simple and achieves high quality results with less training due to a smaller number of hyper parameters.

Our method makes the following technical contributions:
\begin{itemize}
    \item We develop a novel network architecture that combines vector neurons and self attention enabling rotational invariance. The network is small in the number of hyper-parameters
    \item We introduce a novel multi-headed self attention layer (VNAttention) that incorporates vector neurons through an adaptation that is simple to implement.
    \item In this context we define novel feed forward linear layers that operate on vector representations (VNLinear)
\end{itemize}

\section{Related Work}

2D images are efficiently processed by deep neural networks due to their underlying regular grid. Unlike 2D images, point clouds are essentially unordered and scattered sets in 3D space, making it challenging to design neural networks for them. 
While order invariance has been widely explored for point cloud neural networks, rotational invariance has been just starting to gain popularity.
Our previous work discussion is in the domain of 3D point cloud deep learning and 
focuses on rotation equivariant and invariant network architectures.

Rather than projecting or quantizing irregular 3D point clouds onto regular grids in 2D or 3D, researchers have designed deep network structures that ingest point clouds directly.
PointNet~\cite{qi2017pointnet} utilizes permutation-invariant operators such as pointwise MLPs and pooling layers to aggregate features across a set. PointNet++~\cite{qi2017pointnet2} applies these ideas within a hierarchical spatial structure to increase sensitivity to local geometric layout.
PointCNN~\cite{li2018pointcnn} proposes to reorder the input unordered point clouds with special operators.
Related to us, spherical convolutions were introduced~\cite{esteves2018learning} to address the problem of 3D rotation equivariance and attention has been applied to point graphs convolutions in GACNet~\cite{wang2019graph}.

%Sampling strategies may have significant effect on the model performance and a variety of sampling strategies have been developed~\cite{qi2017pointnet2,dovrat2019learning,wu2019pointconv,yang2019modeling,hu2020randla}.
%%

%Graphs connecting the point set together enable message passing and thus applicable for graph convolution neural networks ~\cite{simonovsky2017dynamic,landrieu2018spg,shen2018mining,wang2018local,wang2019dgcnn}.

\textbf{Transformers and Self-attention}

Transformers  and self-attention were originally introduced for natural
language processing~\citet{vaswani2017}, and have since become the state of the art there~\cite{wu2019pay,Devlin2018,dai2019transformer,Yang2019xlnet}. 
%Transformer-based models often require pre-training on large sets and then fine-tuned for the task at hand~\citep{devlin19-bert,radford2018-gpt,radford2019-gpt2,brown2020-gpt3}.
%
Direct application of self-attention to images would yield quadratic cost in the number of pixels. Thus, local multi-head self attention blocks were introduced to address image convolutions.
Along this path, scalar dot-product self-attention within local image patches were introduced~\cite{hu2019local,ramachandran2019stand}. A family of feature vectors self-attention operators were showed in~\cite{zhao2020san}.

Sparse Transformers were suggested~\citep{child2019-sparsetransformers} to enable scalable approximations to global self-attention.
Similarly, blocks of varying sizes were utilized in~\citep{weissenborn2019-savm} to scale attention.

To allow full self-attention on images, a set of patches are extracted from the image pixels to which all-to-all self attention is applied~\citet{cordonnier2020-sacnn,dosovitskiy2021image}. Nevertheless, patch number is still high, making the model applicable only to small to medium resolution images.
Similarly, Transformers were applied to image pixels after reducing image resolution and color space~\citep{chen20-igpt}. 
%The model is trained in an unsupervised maner as a generative model, and the resulting representation can then be fine-tuned.

Since transformers and self-attention networks outperform convolutional networks on 2D images, our work harness their effective power to point clouds.
In fact, the self-attention mechanism is a dot-product set operator and therefore seems particularly suitable for unordered point sets.
There are a number of previous works that utilize attention for point cloud analysis~\cite{xie2018attentional,liu2019point2sequence,yang2019modeling,lee2019set}.
Nevertheless, they apply attention on the whole point cloud, which introduces heavy computation and renders these approaches inapplicable to large point sets

Recently, local vector self-attention was introduced~\cite{2021hengshuang} which enables scalability to large point sets. Authors use vector attention however their vectors focus on feature representation while our vector neurons focus on rotational equivaiance.

\textbf{Rotation Equivariant and Invariant Networks}

Rotation invariance is a desirable property for tasks like shape classification and segmentation. The lack of robustness to rotation of classical deep learning architectures has driven interest for rotation invariant and equivariant designs.

Many rotation invariant architectures
\cite{liu2018deep, poulenard2019effective, chen2019clusternet, zhang2019rotation, zhang2020global, li2021rotation, zhao2019rotation, rao2019spherical} have been proposed. For example, \cite{chen2019clusternet, zhang2019rotation, zhang2020global, li2021rotation} introduce rotation invariant operations. 
%GC-Conv \cite{zhang2020global} relies on multi-scale reference frames based on PCA. RI-Framework \cite{li2021rotation} and LGR-Net~\cite{zhao2019rotation} pairs local invariant information with global context.

Some works like LGR-Net \cite{zhao2019rotation} use surface normals in addition to the points coordinates. SFCNN \cite{rao2019spherical} proposes an approach similar to multi-view by mapping input pointclouds to a sphere and performing operations on the sphere. Other works like \cite{liu2018deep, poulenard2019effective} rely on more principled approaches borrowing tools from equivariant deep learning.

Recently multiple rotation equivariant deep learning architectures have been showed.
For example, ~\citet{qi2017pointnet} achieved approximate pose equivariance by factoring out SO(3) transformations through object pose estimation.
In general, equivariant networks are built on the theory of $\mathrm{SO}(3)$ representations \cite{thomas2018tensor, kondor2018clebsch, esteves2018learning, weiler20183d, anderson2019cormorant}.
Most of these works rely on the concept of convolution with steerable kernel bases. A steerable kernel basis is a family of functions undergoing a rotation in function/feature space given a rotation of their input~\cite{lang2020wigner}. Thus, features computed through these convolutions are SO(3) equivariant.

Recently, the universality of point cloud rotation equivariance has been studied in \cite{dym2020universality}.
Vector Neurons were introduced in~\cite{deng2021vn} enabling SO(3)-equivariant neural networks. Their work extends neural networks from 1D scalars to
3D vectors which allow mapping of
SO(3) actions to latent spaces through definition of equivariant neural operations. Our work is inspired by Vector Neurons and takes a step further, incorporating them into attention layers to achieve rotational invariance.

%Other works like EMVnet \cite{esteves2019equivariant} consider a multi-view image based representation of the shapes based on renderings of meshes. 

%\citet{qi2017pointnet} achieved approximate pose equivariance by factoring out SO(3) transformations through object pose estimation.
%Most works in the literature study \textit{instance-level} pose estimation, where the ground-truth canonical pose of the 3D CAD models corresponding to the input pointcloud is available~\cite{brachmann2014learning}.

%More recently \citet{wang2019normalized} introduced \textit{category-level} pose estimation, and extension to articulated objects has also been proposed~\cite{li2020categorylevel}.
%While both these methods~\cite{wang2019normalized,li2020categorylevel} need explicit 2D-to-3D supervision, relaxing supervision is possible by borrowing ideas from Transforming~Auto-Encoders~\cite{hinton2011transforming,unsupervised}.
%However, while \citet{sun2020caca} learn \textit{category-level} as well as \textit{multi-category} pose estimation in a fully unsupervised fashion, the underlying equivariant backbone~\cite{sun2020acne} is only equivariant by \textit{augmentation}. 

\section{Method}

We introduce VNT-Net based on a combination between vector neurons ~\cite{deng2021vn} and multi-headed self attention (VNAttention). For this purpose, we define an extension to scaled dot-product attention~\cite{vaswani2017attention} that provides SO(3) equivariance.

In our work, the input is defined as \(\mathcal{V} = \{\vv{V}_1,\vv{V}_2, ... \vv{V}_N\} \in \mathbb{R}^{N \times C \times D}\) where \(N\) is the number points, \(C\) is the number of channels in each point and \(D\) is the size of the VN in each channel (in our case \(D=3\)). \\
Given a weight matrix \(W\in\mathbb{R}^\mathit{C' \times C}\), we define a linear operation:\\
\(\mathit{f}_{lin}(\cdot ;\mathbb{W})\) operating on a vector-list feature
\(\vv{V_i} \in \mathcal{V}\) as follows:
\[\mathit{\vv{V_i}'}=\mathit{f}_{lin}(\vv{V_i} ;\mathbf{W})=\mathbf{W\vv{V_i}}\in\mathbb{R}^{C' \times D}\]

Neurons in standard artificial neural networks are essentially 1D scalars \(z \in \mathbb{R}\). When stacked in a list, these neurons form a \(C^{(d)}\) sized latent feature \(\mathbf{z}=[z_1,z_2, \cdots , z_{C^{(d)}}]^T \in \mathbb{R}^{C^{(d)}}\), where $(d)$ indexes the layer depth.\\

However, when processing un-ordered data embedded in \(\mathbb{R}^3\), like 3D point clouds, realizing the effect of SO(3) rotations applied to the input shape on these vector hidden layers is not obvious.

We are interested in constructing a rotation-equivariant self attention layer that commutes with the action of the SO(3) group.\\
To achieve this, we adapt the neuron representation from a 1D scalar \(z\in \mathbb{R}\) to a 3D vector \(v\in\mathbb{R}^3\), leading to the definition of vector neuron. This results in a list of vector neurons \(\mathbf{V}=[v_1,v_2, \cdots,v_C]^T \in \mathbb{R}^{C\times 3}\) (matrix).

This representation allows us to construct SO(3) equivariat self attention layers and finally an SO(3) invariant transformer network.
% \\
% We define the changes in the number of latent channels \(C^{(d)}\) between layers via the mapping:
% \begin{equation*}
%     \mathcal{V}^{(d+1)} = f(\mathcal{V}^{(d)} ; \theta): \mathbb{R}^{N\times C^{(d)}\times 3} \rightarrow \mathbb{R}^{N \times C^{(d+1)}\times 3}
% \end{equation*}
% where \(\theta\) represents the learnable parameters.
% \\

%Using the VN representation allows us to combine linear layers form VNN \cite{deng2021vn}
For this purpose we construct a self-attention vector neurons (VNAttention) layer and a vector neuron feed-forward network (VN-FFN) that satisfies rotation equivariance , namely, for any rotation matrix \(\mathbf{R}\in SO(3)\):
\begin{equation*}
    \textbf{VNAttention}(\vv{Q}R,\vv{K}R,\vv{V}R)= \textbf{VNAttention}(\vv{Q},\vv{K},\vv{V})R
\end{equation*}
\begin{equation*}
    \textbf{VN-FFN}(\vv{V}R) = \textbf{VN-FFN}(\vv{V})R
\end{equation*}
\\

\begin{figure}[t]
    \begin{minipage}{0.5\textwidth}%
        \centering
        \includegraphics[width=0.6\linewidth]{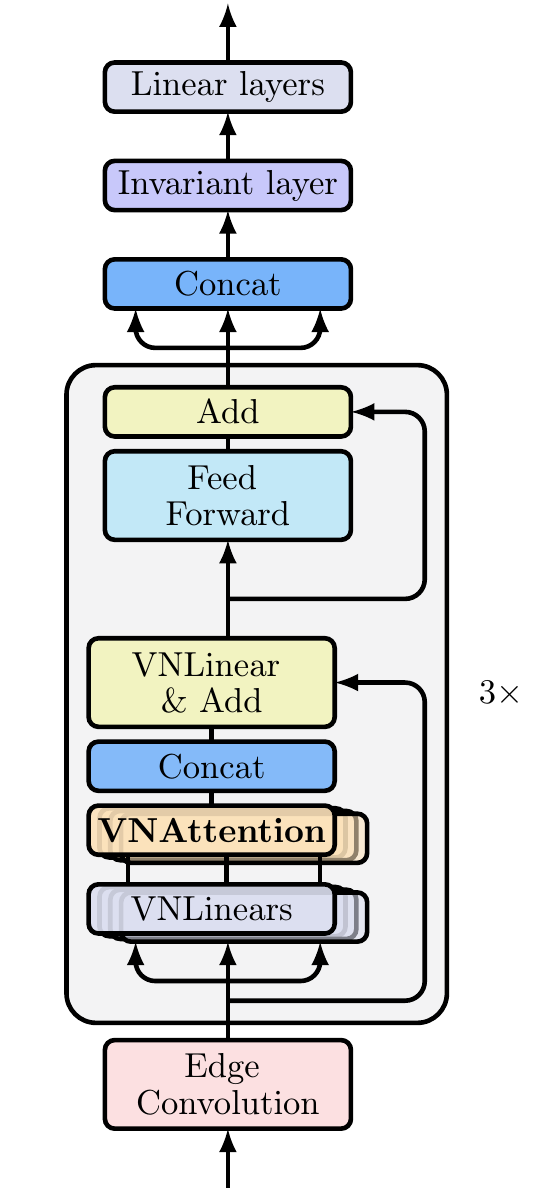}
    \end{minipage}%
    \begin{minipage}{0.5\textwidth}
        \centering
        \includegraphics[width=0.7\linewidth]{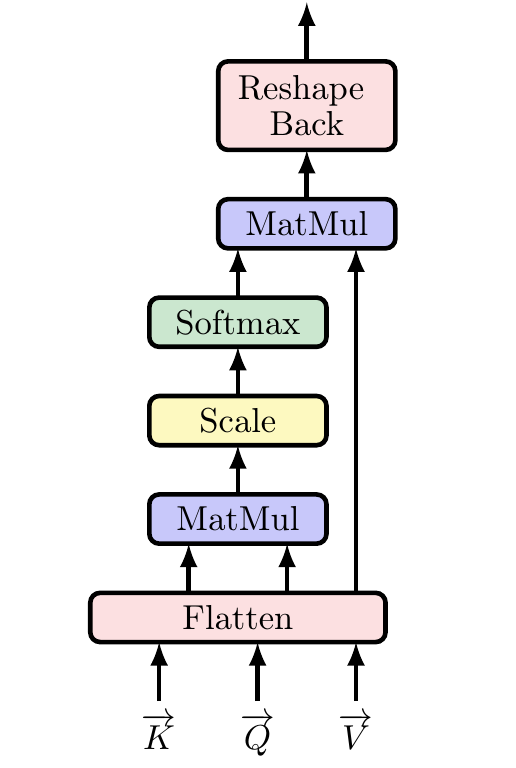}
    \end{minipage}%
    \caption{VNT-Network architecture (left) and VNAttention layer structure (right).}
    \label{fig:dotprod}
\end{figure}

% \textbf{VNAttention Layer}

Self-attention is the central layer of any transformer network. We base our VNAttention layer on scaled dot-product attention~\cite{vaswani2017attention}(see Figure \ref{fig:dotprod} right):
\begin{gather*}
    \textbf{Attention}(Q,K,V) = \textbf{softmax}(\frac{QK^T}{\sqrt{d_k}})V \\
    W = \textbf{softmax}(\frac{S}{\sqrt{d_k}}) = \textbf{softmax}(\frac{QK^T}{\sqrt{d_k}})
\end{gather*}
In the case of vector neurons, we define the parameters in the following way:
\begin{equation*}
    \vv{Q},\vv{K},\vv{V} \in \mathbb{R}^{N \times C \times D} , ~~ d_k ~~ \text{is the channel size of Q and K.}
\end{equation*}
We define \(\vv{X}_{(c)}\) as the indexing of a channel \(c\) in \(\vv{X}\) (where \(\vv{X}\in \{\vv{Q},\vv{K},\vv{V}\}\) and \(c \in \{1,...,C\}\)).\\
% \begin{equation*}
%     X_{(c)} = \{x \in X, x \in \mathbb{R}^{C\times D}| x_c \in \mathbb{R}^D \} \in \mathbb{R}^{N\times D}
% \end{equation*}
Finally, we define VNAttention as follows:
\begin{equation*}
    \textbf{VNAttention}(\vv{Q},\vv{K},\vv{V})=\textbf{softmax}(\frac{\sum_{c \in C}{\vv{Q}_{(c)} \vv{K}_{(c)}^T}}{\sqrt{d_k}})\vv{V}
\end{equation*}

For simplicity, we easily show that in the case where $D = 1$ \textbf{VNAttention} boils down to the original \textbf{Attention}:
\begin{gather*}
    \textbf{VNAttention}(\vv{Q},\vv{K},\vv{V})=\textbf{softmax}(\frac{\sum_{c \in C}{\vv{Q}_{(c)} \vv{K}_{(c)}^T}}{\sqrt{d_k}})\vv{V} \\ =
    \textbf{softmax}(\frac{{\begin{bmatrix}
                \vv{Q}_1, \cdots, \vv{Q}_C
            \end{bmatrix}  \begin{bmatrix}
                \vv{K}_1, \cdots, \vv{K}_C
            \end{bmatrix}^T}}{\sqrt{d_k}})\vv{V} = \textbf{softmax}(\frac{{Q K^T}}{\sqrt{d_k}})V = \textbf{Attention}(Q,K,V)
\end{gather*}

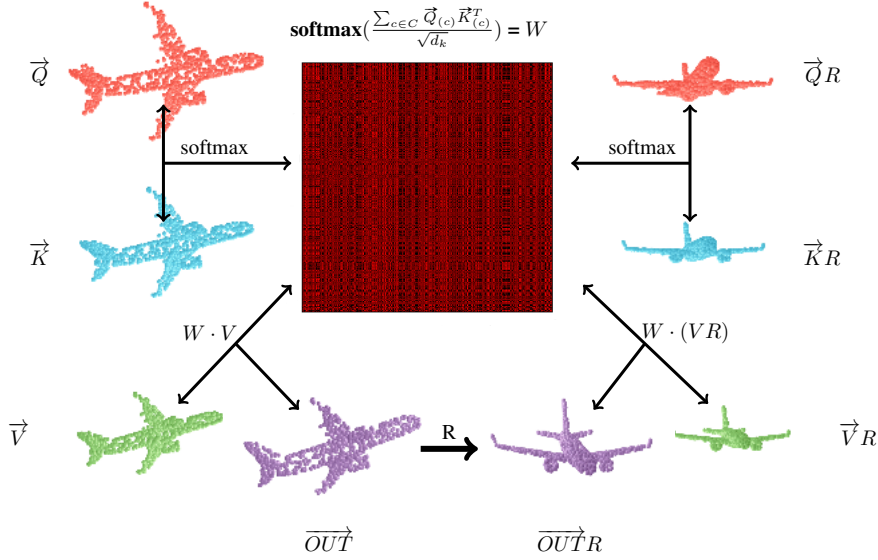
\begin{figure}[t]
    \centering
    \scalebox{0.8}{%
    \begin{tikzpicture}
        \node[inner sep=0pt, label=above:\(\textbf{softmax}(\frac{\sum_{c \in C}{\vv{Q}_{(c)} \vv{K}_{(c)}^T}}{\sqrt{d_k}}) ~ \textbf{=} ~ W\)] (mat) at (4.5,-2)
        {\includegraphics[width=.35\textwidth]{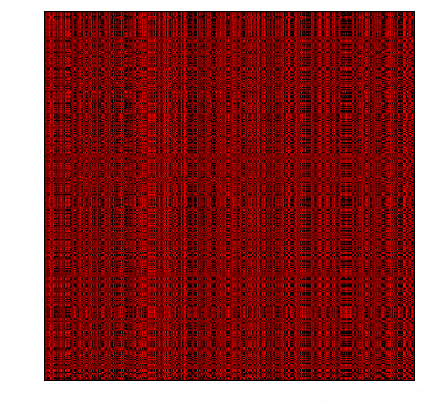}};
        \node[inner sep=0pt, label=left:\(\overrightarrow{Q}\)] (Qleft) at (0.25,0)
        {\includegraphics[width=.25\textwidth]{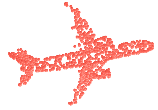}};
        \node[inner sep=0pt, label=right:\(\overrightarrow{Q}R\)] (Qright) at (9,0)
        {\includegraphics[width=.25\textwidth]{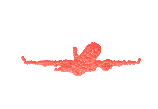}};

        \node[inner sep=0pt, label=left:\(\overrightarrow{K}\)] (Kleft) at (0.25,-3)
        {\includegraphics[width=.25\textwidth]{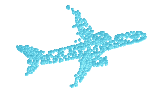}};
        \node[inner sep=0pt, label=right:\(\overrightarrow{K}R\)] (Kright) at (9,-3)
        {\includegraphics[width=.25\textwidth]{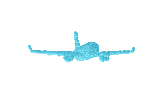}};

        \node[inner sep=0pt, label=left:\(\overrightarrow{V}\)] (Vleft) at (-.1,-6)
        {\includegraphics[width=.25\textwidth]{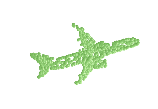}};
        \node[inner sep=0pt, label=right:\(\overrightarrow{V}R\)] (Vright) at (9.6,-6)
        {\includegraphics[width=.25\textwidth]{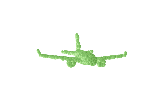}};

        \node[inner sep=0pt, label=below:\(\overrightarrow{OUT}\)] (outleft) at (3,-6.25)
        {\includegraphics[width=.25\textwidth]{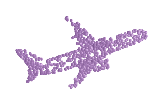}};
        \node[inner sep=0pt, label=below:\(\overrightarrow{OUT}R\)] (outright) at (7,-6.25)
        {\includegraphics[width=.25\textwidth]{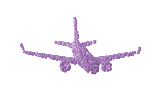}};
        \node[inner sep=0pt] (outleftancor) at (4.5,-6.25) {};
        \node[inner sep=0pt] (outrightancor) at (5.5,-6.25) {};
        \draw[-> ,line width=1mm] (outleftancor) -- node [above] {R} (outrightancor);

        \node[inner sep=0pt] (Qleftancher) at (0.25,-0.5) {};
        \node[inner sep=0pt] (Kleftancher) at (0.25,-2.5) {};
        \draw[<-> ,line width=0.5mm] (Qleftancher) -- (Kleftancher);
        \node[inner sep=0pt] (Qrightancher) at (9,-0.5) {};
        \node[inner sep=0pt] (Krightancher) at (9,-2.5) {};
        \draw[<-> ,line width=0.5mm] (Qrightancher) -- (Krightancher);
        \node[inner sep=0pt] (leftsidemat) at (2.4,-3.5) {};
        \node[inner sep=0pt] (Vleftplace) at (0.5,-5.5) {};
        \node[inner sep=0pt] (midVmatleft) at (1.45,-4.5) {};
        \node[inner sep=0pt] (rightsidemat) at (7.2,-3.5) {};
        \node[inner sep=0pt] (Vrightplace) at (9.3,-5.5) {};
        \node[inner sep=0pt] (midVmatright) at (8.25,-4.5) {};

        \node[inner sep=0pt] (leftqkcenter) at (0.25,-1.5) {};
        \node[inner sep=0pt] (matleft) at (2.4,-1.5) {};
        \node[inner sep=0pt] (rightqkcenter) at (9,-1.5) {};
        \node[inner sep=0pt] (matright) at (7,-1.5) {};

        \draw[<-> ,line width=0.5mm] (Vleftplace.center) -- node [left, pos=0.6] {$W\cdot V$} (leftsidemat);
        \draw[<-> ,line width=0.5mm] (Vrightplace.center) -- node [right, pos=0.6] {$W \cdot (VR)$} (rightsidemat);

        \draw[-> ,line width=0.5mm] (leftqkcenter.center) -- node [above, pos=0.4] {softmax} (matleft);
        \draw[-> ,line width=0.5mm] (rightqkcenter.center) -- node [above, pos=0.4] {softmax} (matright);

        \node[inner sep=0pt] (outleftarrow) at (2.5,-5.6) {};
        \node[inner sep=0pt] (outrightarrow) at (7.4,-5.6) {};
        \draw[-> ,line width=0.5mm] (midVmatleft.center) -- (outleftarrow.center);
        \draw[-> ,line width=0.5mm] (midVmatright.center) -- (outrightarrow.center);
    \end{tikzpicture}
    }

    \caption{VNAttention attention matrix rotation invariance demonstration.}
        % This is a draw of the Proof from \ref{proof}
        % This show how the VNAttention layer is equivariance to the input. On the left we see the input before we Rotation (RED $\protect\overrightarrow{Q}$, Blue $\protect\overrightarrow{K}$, Green $\protect\overrightarrow{V}$ ). On the right we see the input after Rotation (RED $\protect\overrightarrow{Q}R$, Blue $\protect\overrightarrow{K}R$, Green $\protect\overrightarrow{V}R$ ). On the middle we see the matrix produce by \(\text{softmax}(\frac{\sum_{c \in C}{\protect\vv{Q}_{(c)} \protect\vv{K}_{(c)}^T}}{\sqrt{d_k}})\)  matrix (Red Heat map)., we get the same result for rotated and not rotated. On the bottom we see the output ($\protect\overrightarrow{OUT}$ and $\protect\overrightarrow{OUT}R$ Purple) .
    \label{fig:prof}
\end{figure}

Figure~\ref{fig:prof} demonstrates the equivariance and invariance to the rotation group of the attention matrix (red square). Given an input (left airplane) and a rotated one (right airplane), the attention layer structure addresses rotation as commutative (bottom part) and thus is equivariant to rotation group.

\textbf{Flattening}

In order to optimize the VNAttention layer, we flatten its vector neurons input. This allows performing a single matrix multiplication instead of numerous dot products as summation of row-col multiplications.
Thus we flatten the $\vv{Q}$ and $\vv{K}$ matrices as follows:
\begin{gather*}
    \vv{Q}, \vv{K} \in \mathbb{R}^{N \times C \times D}\\
    \vv{Q}_{flat} = [\vv{Q}_{(0)},\vv{Q}_{(1)},\cdots ,\vv{Q}_{(C)}], \vv{K}_{flat} = [\vv{K}_{(0)},\vv{K}_{(1)},\cdots ,\vv{K}_{(C)}] \\
    \vv{Q}_{flat},\vv{K}_{flat} \in \mathbb{R}^{N \times CD}\\
    \vv{Q}_{flat} \cdot \vv{K}_{flat}^T = [\vv{Q}_{(0)},\vv{Q}_{(1)},\cdots ,\vv{Q}_{(C)}] \cdot  [\vv{K}_{(0)},\vv{K}_{(1)},\cdots ,\vv{K}_{(C)}]^T \\
    = \sum_{c \in C}{\vv{Q}_{(c)} \vv{K}_{(c)}^T}
\end{gather*}
In this way we don't need to save intermediate representations of \(S_{(c)}\).

\textbf{Multi-Headed VNAttention layers}

As is the common practice in Transformer networks, we found that it is beneficial to use multi-headed attention, instead of performing a single attention function with \(d_{model}\)-dimensional keys, values and queries.

We achieve this by linearly projecting the queries, keys and values \textit{h} times with different learned linear projections to \(d_k\) and \(d_v\) dimensions, respectively. On each of these projected versions of queries, keys and values we then perform the attention function in parallel, yielding \(d_v\)-dimensional output values. These are concatenated and once again projected.\\
The following equations describe this process mathematically:
\begin{gather*}
    W_i^Q\in \mathbb{R}^{d_{\text{model}} \times d_k}, W_i^K\in \mathbb{R}^{d_{\text{model}} \times d_k}, W_i^V\in \mathbb{R}^{d_{\text{model}} \times d_v} ,W^O\in \mathbb{R}^{h_dv\times d_{\text{model}}} \\
    head_i  = \textbf{VNAttention}(QW_i^Q,KW_i^K,VW_i^V) \\
    \textbf{Multi-Headed}(Q,K,V) = \textbf{Concat}(head_1,\cdots,head_h)W^O + \mathcal{V}
\end{gather*}

\textbf{VN Feed-Forward layer}

We found out that using a feed forward layer is beneficial after the multi-headed layers.
Thus, after each multi-headed VNAttention block we add a VN Feed-Forward layer as defined in~\cite{deng2021vn}.
\begin{equation*}
    \textbf{VN-FFN} (\vv{V}) = f_{lin}(\textbf{VN-ReLU}(f_{lin}(\vv{V};W_1));W_2) = \textbf{VN-ReLU}(W_1\vv{V})W_2
\end{equation*}

\textbf{Vector Transformer Network}

The basic building block of our network is constructed as follows:
\textbf{Multi-Headed VNAttention} block, Residual connection,  \textbf{VN Feed-Forward} block, and Residual connection.
We construct our network from 3 stacks of basic blocks, that is, the input for each block is the output of the previous block. After which, we concatenate the output of the 3 blocks, then we pass it to an Invariant Layer, similar to the layer used on VNN \cite{deng2021vn}, to get an SO(3) invariant representation of the output. The last part of the network is a simple classification head to get scores (see Figure~\ref{fig:dotprod} (left figure)).\\

For part-segmentation setups, we use the same architecture with 2 small changes.
We pass the categorical data through a simple linear layer, then concatenate it's output to the output of the Invariant Layer.
We change replace the classification head with a simple segmentation head.

\textbf{Linear Dependency Problem}

Similar to the observation in ~\cite{deng2021vn}, when encoding data across network layers, thus resizing it, vectors may become linearly dependent due to their duplication. As a consequence, the output of a linear layer applied to them would degenerate to a set of linearly dependent vectors n \(\mathbb{R}^{C\times 3}\) (pointing to the same direction). This is analogous to applying a per-pixel 1x1 convolution to a gray-scale image (single input channel).

To deal with this issue, we add to our architecture an edge convolution block in the input layer, mapping features from \(\mathbb{R}^{1\times 3}\) to features in \(\mathbb{R}^{C\times 3}\) with \(C > 1\), and those are put back to the basic block stack.

\section{Results}
We evaluate our model on two core tasks in point cloud processing: classification and segmentation. This shows that our network works well on diverse types of tasks. In this section we discuss our datasets, network implementation and performance, result and comparisons.

\textbf{Datasets}

We employed the ModelNet40 \cite{wu2015modelnet}
and ShapeNet \cite{wu2015modelnet} datasets for evaluation. The ModelNet40 dataset consists of 40 classes with 12,311 CAD models in total. In the classification setup, we used 9,843 CAD models for training and the rest for testing.

For segmentation on the ShapeNet dataset, we followed \cite{yi2016scalable} by using ShapeNet-part for part segmentation, which has 16 shape categories with more than 30,000 models.

\textbf{Rotated Dataset}
In both classification and segmentation, we train and test our network on both the original dataset as well as its rotated versions denoted as follows: z/z, z/SO(3), and SO(3)/SO(3). z stands for data augmentation with rotations only around the z-axis and SO(3) stands for arbitrary rotations. All rotations are generated randomly at training time.

The notation A/B denotes the training data was A and the testing data was B. The aim here was to evaluate the capability of our network as well as SOTA networks to train and test on similar sets z/z and SO(3)/SO(3) and to generalize on z/SO(3).

\textbf{Data Augmentation.}

In the implementation of our network, we found that not all common data augmentation technics are useful. In this section, we will describe what augmentations were useful and what didn't work.

\paragraph{Sampling.}
During the training phase on Modelnet40, we used run farther point sampling on the training data. Where we sample 1024 points on every model randomly each time. we found this help us to avoid overfitting the training set. The reason we found it more useful than just random sampling was that this sampling creates more variety of vectors directions. This will make the attention matrix to be more sparse and find more diverse connections.

Randomly dropping points by changing their value of them to zero.
This augmentation can be found used in most of the implementation of the comparable network. This augmentation didn't help and in 't help and most of the time caused our implementation to not converge.

\paragraph{Normalization.}
Using data that has its scale normalizes gives us much more stable training. In a case, where scale normalization was not utilized the network frequently didn't converge or in the middle of the training stop converging or even got much worst on the test set. The normalization is done by subsubsection of the centroid of the point cloud and dividing by the maximum vector length from all the points inside the point cloud.

\paragraph{Scale.}
Random scaling augmentation is useful in the range of 0.8 to 1.25. This is even when the \textbf{VNAttention} attention matrix is not affected by the size scaling because of the softmax operation.

\paragraph{Shift.}
 Random Point shifting was useful to add noise to our network and avoid overfitting. We randomly shift all our points in the range of -0.1 to 0.1.

\textbf{Network implementation}

The principal hyper-parameters of our network are as follows:
\begin{enumerate}
    \item \textbf{Linear Dimension}: The feature dimension of the output of the edge convolution, as well as that of the input/output of each MultiHeaded VNAttention and Feed Forward layer.
    \item \textbf{Heads Number}: The numbers of heads in each MultiHeaded VNAttention layer.
    \item \textbf{Head Size}: The feature dimension of each head in each MultiHeaded VNAttention layer.
\end{enumerate}
For classification, the setup that worked best for us is: \textbf{Linear Dimension}=16, \textbf{Heads Number}=24, \textbf{Head Size}=16.
For segmentation, the setup that worked best for us is: \textbf{Linear Dimension}=128, \textbf{Heads Number}=14 and \textbf{Head Size}=16.

\textbf{Classification Evaluation and Comparison}

We evaluate classification results on ModelNet40 comparing with VN networks \cite{deng2021vn}, and SOTA point transformer networks \cite{2021hengshuang} \cite{2021PCT} and SOTA rotation invariant networks (see Table \ref{tab:exp:cls}).

The table is divided into sections top to bottom as follows. First are vector neurons networks and DGCNN, next are point cloud convolution networks, following are rotational-sensitive and invariant networks, transformers and finally networks using points+normals representations as input.

The highest accuracy network for z/z is Shellnet~\cite{zhang2019shellnet} however not rotational invariant. For z/SO(3) and SO(3)/SO(3) highest accuracy network is LGR-Net however using a very large number of hyper-parameters (1.37M vs. 5.55M).

Nevertheless, our network is 2nd place in z/SO(3) and third place in SO(3)/SO(3) and 7th place in z/z

Note that although data augmentation is efficient improving the majority of networks performance (compare  z/SO(3) and SO(3)/SO(3) columns), our method still achieved better results in both cases.

\textbf{Semantic Segmentation Evaluation and Comparison}

We evaluate semantic segmentation results on the ShapeNet part-segmentation in average category mean IoU~\ref{tab:exp:seg} and global mean IoU~\ref{tab:exp:rotations_seg} where Training is done on aligned data without rotation augmentation.

The table is divided into sections top to bottom as follows. First are vector neurons networks and DGCNN, next are point cloud convolution networks, following are rotational-sensitive and invariant networks and finally transformers.

The highest accuracy network For z/SO(3) and SO(3)/SO(3) highest accuracy network is VN-DGCNN~\cite{deng2021vn} however it have less accuracy over the global mean IoU (84.4 vs 81.5).

Nevertheless, our network is 2nd place in z/SO(3) and SO(3)/SO3(3) in average category mean IoU.
In the case of global mean IoU our network achieves first place in I/z and I/SO(3) and 2nd place I/I however The first place is not invariant.

\begin{table}[h!]
\centering
% \resizebox{.5\textwidth}{!}{%
\begin{tabular}{c c c c}
\hline
Methods                                                   & z/z           & z/SO(3)       & SO(3)/SO(3)   \\ \hline
% PointNet \cite{qi2017pointnet}           & 85.9          & 19.6          & 74.7          \\
DGCNN \cite{wang2019dynamic}             & 90.3          & 33.8          & 88.6          \\
% VN-PointNet \cite{deng2021vn}            & 77.5          & 77.5          & 77.2          \\
VN-DGCNN    \cite{deng2021vn}            & 89.5          & 89.5          & 90.2          \\ \hline
PCNN \cite{atzmon2018point}              & 92.3          & 11.9          & 85.1          \\
ShellNet \cite{zhang2019shellnet}        & \textbf{93.1}          & 19.9          & 87.8          \\
PointNet++ \cite{qi2017pointnetpp}       & 91.8          & 28.4          & 85.0          \\
PointCNN \cite{li2018pointcnn}           & 92.5          & 41.2          & 84.5          \\
Spherical-CNN \cite{esteves2018learning} & 88.9          & 76.7          & 86.9          \\
$a^3$S-CNN \cite{liu2018deep}            & 89.6          & 87.9          & 88.7          \\ \hline
SFCNN \cite{rao2019spherical}            & 91.4          & 84.8          & 90.1          \\
TFN \cite{thomas2018tensor}              & 88.5          & 85.3          & 87.6          \\
RI-Conv \cite{zhang2019rotation}         & 86.5          & 86.4          & 86.4          \\
SPHNet \cite{poulenard2019effective}     & 87.7          & 86.6          & 87.6          \\
ClusterNet \cite{chen2019clusternet}     & 87.1          & 87.1          & 87.1          \\
GC-Conv \cite{zhang2020global}           & 89.0          & 89.1          & 89.2          \\
RI-Framework \cite{li2021rotation}       & 89.4          & 89.4          & 89.3          \\ \hline
\textbf{VNT}                             & 89.2          & 89.2          & 88.6          \\
\textbf{VNT+N} & 90.3          & 90.3          & 90.3 \\
Point Transformer \cite{2021hengshuang}  & 89.0          & 31.0          & 86.1          \\
PCT \cite{2021PCT}                       & 90.0          & 28.3          & 87.0          \\ \hline
SFCNN \cite{rao2019spherical}            & 92.3          & 85.3          & 91.0          \\
LGR-Net \cite{zhao2019rotation}          & 90.9          & \textbf{90.9}          & \textbf{91.1}          \\ \hline
\end{tabular}%
% }
\caption{Classification accuracy comparison on the ModelNet40 dataset.}
\label{tab:exp:cls}
\end{table}

\begin{table}[h!]
\centering
% \resizebox{.5\textwidth}{!}{%
\begin{tabular}{ccc}
\hline
Methods                                & z/SO(3)        & SO(3)/SO(3)   \\ \hline
% PointNet \cite{qi2017pointnet}         & 38.0           & 62.3          \\
DGCNN \cite{wang2019dynamic}           & 49.3           & 78.6          \\
% VN-PointNet \cite{deng2021vn}          & 72.4           & 72.8          \\
VN-DGCNN \cite{deng2021vn}             & \textbf{81.4}     & \textbf{81.4}         \\ \hline
PointCNN \cite{li2018pointcnn}         & 34.7           & 71.4          \\
PointNet++ \cite{qi2017pointnetpp}     & 48.3           & 76.7          \\
ShellNet \cite{zhang2019shellnet}      & 47.2           & 77.1          \\ \hline
RI-Conv \cite{zhang2019rotation}       & 75.3           & 75.3          \\
TFN \cite{thomas2018tensor}            & 76.8           & 76.2          \\
GC-Conv \cite{zhang2020global}         & 77.2           & 77.3          \\
RI-Framework \cite{li2021rotation}     & 79.2           & 79.4          \\
LGR-Net\cite{zhao2019rotation}         & 80.0           & 80.1          \\ \hline
\textbf{VNT+N}                         & 80.5           & 80.5          \\
Point Transformer\cite{2021hengshuang} & 70.3          & 40.8          \\
PCT\cite{2021PCT}                      & 77.4          & 43.5
\end{tabular}%
% }
\caption{Semantic-segmentation comparison on ShapeNet.}
\label{tab:exp:seg}
\end{table}

\begin{table}[h!]
\small
\centering
\begin{tabular}{c|ccc}
\hline
Method         & I/I           & I/z           & I/SO(3)       \\ \hline
% PointNet       & 78.7          & 36.7          & 30.3          \\
DGCNN \cite{wang2019dynamic}         & \textbf{85.2}          & 43.8          & 36.1          \\
% VN-PointNet    & 73.0          & 73.0          & 73.0          \\
VN-DGCNN  \cite{deng2021vn}      & 81.5          & 81.5          & 81.5          \\
\textbf{VNT}   & 83.4          & 83.4          & 83.4          \\
\textbf{VNT+N} & 84.4          & \textbf{84.4} & \textbf{84.4} \\ \hline
\end{tabular}%
\caption{Semantic-segmentation comparison on ShapeNet (mIoU).}
\label{tab:exp:rotations_seg}
\end{table}

\begin{table}[h!]
\small
    \centering
    \begin{tabular}{c|c}
    \hline
    Method         & Parameters  \\ \hline
    VN-DGCNN   \cite{deng2021vn}    &  2.90M \\
    LGR-Net  \cite{zhao2019rotation}      &  5.55M \\
    PCT  \cite{2021PCT}          &  2.88M \\
    \textbf{VNT+N} & \textbf{1.37M} \\
    \hline
    \end{tabular}%
    \caption{Hyper-params networks comparisons}
    \label{tab:exp:cls_parms}
    \end{table}

\textbf{Hyper-params Comparison}

We compare the hyper-parameter size of our network w.r.t. SOAT networks in table~\ref{tab:exp:cls_parms}. Our network is significantly smaller than other SOTA networks that perform similarly in terms of accuracy. This is meaningful since a smaller number of parameters means shorter training time and smaller training data sets.

\section{Conclusions}

In this work we combine recently introduced vector neurons with self attention layers to build a point cloud vector neuron transformer network (VNT-Net) that is rotational invariant. 
For this purpose, we develop a novel multi-headed self attention layer (VNAttention) that incorporates vector neurons through an adaptation that is simple to implement.
The network is small in the number of hyper-parameters and results show high accuracy that is competitive with SOTA networks that are significantly larger in terms of number of hyperparams. 
Although our network does not achieve the highest accuracy it provides an interesting set of simple tools to move transformers into rotational invariant and equivariant domains. 
In future work, we plan to take this approach to higher dimensions and adapt vector neurons to new transformer networks. 

%%
%% The next two lines define the bibliography style to be used, and
%% the bibliography file.
\bibliographystyle{plainnat}
\bibliography{ref}

% \input{checklist.tex}
%%
%% If your work has an appendix, this is the place to put it.
\appendix

\section{VNAttention Proof and Analysis}
\textbf{Proof of VNAttention rotational Equivariance}

We will show that \(\mathit{R} \in SO(N)\) commutes with the VNAttention layer (Equivariance):
\begin{gather*}
    \textbf{VNAttention}(\vv{Q}R,\vv{K}R,\vv{V}R)=\textbf{softmax}(\frac{\sum_{c \in C}{\vv{Q}_{(c)}R (\vv{K}_{(c)}R)^T}}{\sqrt{d_k}})\vv{V}R=\\=
    \textbf{softmax}(\frac{\sum_{c \in C}{\vv{Q}_{(c)}R R^T \vv{K}_{(c)}^T}}{\sqrt{d_k}})\vv{V}R=\\=\textbf{softmax}(\frac{\sum_{c \in C}{\vv{Q}_{(c)} \vv{K}_{(c)}^T}}{\sqrt{d_k}})\vv{V}R=\textbf{VNAttention}(\vv{Q},\vv{K},\vv{V})R
\end{gather*}

\textbf{Analysis of the attention weights}

The result of the dot product of \(\vv{Q}_{(c)} \cdot \vv{K}_{(c)}^T\) can also be represented in the \(\textbf{VNAttention}(\vv{Q},\vv{K},\vv{V})\) following way:
\begin{gather*}
    \vv{Q}_{(c)} \in \mathbb{R}^{N \times D},\vv{K}_{(c)} \in \mathbb{R}^{N \times D}\\
    \vv{Q}_{(c)} \cdot \vv{K}_{(c)}^T =
    \begin{bmatrix}
        q_1    \\
        q_2    \\
        \vdots \\
        q_n
    \end{bmatrix} [k_1^T,k_2^T,... k_n^T]
    =\begin{bmatrix}
        s_{11} & \cdots & s_{1n} \\
        \vdots & \ddots & \vdots \\
        s_{n1} & \cdots & s_{nn}
    \end{bmatrix} = S_{(c)}\\
    s_{ij} = q_i\cdot k_j^T = |q_i||k_j|\cos(\theta)
\end{gather*}
For the Matrix \(S\) we get the following equation:
\begin{equation*}
    S_{ij} = \sum_{c \in C}{q_{(c),i}\cdot k_{(c),j}^T = \sum_{c \in C}{|q_{(c),i}||k_{(c),j}|\cos(\theta)}}
\end{equation*}
This equation show that the closer the VNs directions, the larger the weight value will be.
This is very similar to classic self attention, where the more correlated the features are the larger the weight will be.

\section{Classification and Segmentation Heads}

\textbf{Classification Head}.
The classification head is built form 3 linear layers, and we add drop out and leaky-ReLU between the layers. Dropout is set to 50\%.

\textbf{Segmentation Head}.
The segmentation head is built from 4 linear layers, and between each layer we add ReLU and batch-norm.

\section{Extra Hyper Parameters}

Here will describe extra hyper Parameters that where used in the training of the model.

\textbf{Learning Rate and Optimizer}.
We found that using Adam Optimizer with learning rates in the range of 1e-3 to 1e-4 achieves good result. The best results on segmentation and classification were achieved using a learning rate of 5e-4.

\textbf{Batch Size}.
We used a batch size of 8 on GPU of GTX 1080Ti and 16 when using RTX 3090.

\textbf{Scheduler}.
We use learning rate step scheduler with step size of 20 and $\gamma=0.9$.

\end{document}